\documentclass[conference]{IEEEtran}
\usepackage{cite}
\usepackage{amsmath,amssymb,amsfonts}
\usepackage{graphicx}
\usepackage{textcomp}
\usepackage{xcolor}
\usepackage{algorithm}
\usepackage{algpseudocode}
\usepackage{color, colortbl}
\usepackage{amsmath}
\usepackage{subfig}
\usepackage{algorithmicx}

\makeatletter
\newcommand{\newlineauthors}{%
  \end{@IEEEauthorhalign}\hfill\mbox{}\par
  \mbox{}\hfill\begin{@IEEEauthorhalign}
}
\makeatother
\definecolor{Gray}{gray}{0.9}
\def\BibTeX{{\rm B\kern-.05em{\sc i\kern-.025em b}\kern-.08em
    T\kern-.1667em\lower.7ex\hbox{E}\kern-.125emX}}
\begin{document}
\setlength{\belowcaptionskip}{-3pt}

\title{BROAD AREA SEARCH AND DETECTION OF SURFACE-TO-AIR MISSILE SITES USING SPATIAL FUSION OF COMPONENT 
OBJECT DETECTIONS FROM DEEP NEURAL NETWORKS}

\author{\IEEEauthorblockN{Alan B. Cannaday II}
\IEEEauthorblockA{\textit{Center for Geospatial Intelligence} \\
\textit{University of Missouri}\\
Columbia, MO, USA \\
abcannaday@mail.missouri.edu}
\and
\IEEEauthorblockN{ Curt H. Davis}
\IEEEauthorblockA{\textit{Center for Geospatial Intelligence} \\
\textit{University of Missouri}\\
Columbia, MO, USA \\
DavisCH@missouri.edu}
\and
\IEEEauthorblockN{Grant J. Scott}
\IEEEauthorblockA{\textit{Center for Geospatial Intelligence} \\
\textit{University of Missouri}\\
Columbia, MO, USA \\
GrantScott@missouri.edu}
\newlineauthors
\IEEEauthorblockN{Blake Ruprecht}
\IEEEauthorblockA{\textit{Mizzou INformation and Data FUsion Laboratory} \\
\textit{University of Missouri}\\
Columbia, MO, USA \\
bcrf53@mail.missouri.edu}
\and
\IEEEauthorblockN{Derek T. Anderson}
\IEEEauthorblockA{\textit{Mizzou INformation and Data FUsion Laboratory} \\
\textit{University of Missouri}\\
Columbia, MO, USA \\
andersondt@missouri.edu}
}

\maketitle

\begin{abstract}
Here we demonstrate how Deep Neural Network (DNN) detections of multiple constitutive or component objects that are part of a larger, more complex, and encompassing feature can be spatially fused to improve the search, detection, and retrieval (ranking) of the larger complex feature. First, scores computed from a spatial clustering algorithm are normalized to a reference space so that they are independent of image resolution and DNN input chip size. Then, multi-scale DNN detections from various component objects are fused to improve the detection and retrieval of DNN detections of a larger complex feature. We demonstrate the utility of this approach for broad area search and detection of Surface-to-Air Missile (SAM) sites that have a very low occurrence rate (only 16 sites) over a $\sim$90,000 km\textsuperscript{2} study area in SE China. The results demonstrate that spatial fusion of multi-scale component-object DNN detections can reduce the detection error rate of \textit{SAM Sites} by $>$85\% while still maintaining a 100\% recall. The novel spatial fusion approach demonstrated here can be easily extended to a wide variety of other challenging object search and detection problems in large-scale remote sensing image datasets.
\end{abstract}

\section{Introduction}
Within the last five years, Deep Neural Networks (DNN) have shown through extensive experimental validation to deliver outstanding performance for object detection/recognition in a variety of benchmark high-resolution remote sensing image datasets \cite{b1}-\hspace{1sp}\cite{b13}. Methods such as You Only Look Once (YOLO) \cite{yolo}, region-based CNN (R-CNN) \cite{rcnn}, and derivations thereof \cite{fastrcnn}-\hspace{1sp}\cite{Koga} have all shown promising results for a variety of object detection applications in remote sensing imagery.

The demonstrated ability of DNNs to automatically detect a wide variety of man-made objects with very high accuracy has tremendous potential to assist human analysts in labor-intensive visual searches for objects of interest in high-resolution imagery over large areas of the Earth’s surface. However, the vast majority of published studies for DNN object detection in remote sensing imagery have focused on development of new deep learning algorithms/methods and/or comparative testing/evaluation of these methods on benchmark datasets (both public and private). 

As noted by Xin \textit{et al.} \cite{xin}, comparatively fewer studies have attempted to apply promising DNN methods to demonstrate efficacy and/or further develop these new methods via applications to large-scale or broad area remote sensing image datasets, e.g. \cite{delmarco}-\hspace{1sp}\cite{b3}. Since “large-scale” or “broad area” are subjective descriptors, here we define these to be applications where the algorithm is applied to validation image datasets, i.e. excluding training data, covering an area greater than 1,000 km\textsuperscript{2}. 

Further, even DNN detectors that demonstrate exceptionally high accuracy (e.g. 99\%) on benchmark testing datasets will still generate a tremendous number of errors when applied to large-scale/broad area remote-sensing image datasets. For example, a DNN detector with 99\% average accuracy, chip size of 128 x 128 pixels, and a chip scan overlap of 50\% will generate 88,000 errors when applied to a 0.5 m GSD image dataset covering an area of interest (AOI) of 10,000 km\textsuperscript{2} (e.g. 1\textsuperscript{o} x 1\textsuperscript{o} cell). 

If post DNN detection results are intended to be reviewed by human analysts in machine-assisted analytic workflows, then large numbers of detection errors can quickly lead to “error fatigue” and a corresponding negative end-user perception of machine-assisted workflows. Thus, it is important to develop methods to reduce error rates resulting from application of DNN detectors to large-scale/broad area remote-sensing image datasets to improve machine-assisted analytic workflows.

In this study we develop a new framework for spatially fusing multi-scale detections from a variety of component objects to improve the detection and retrieval of a larger complex feature.  A key aspect of this framework is the development of a spatial clustering algorithm that generates normalized per-object cluster scores to facilitate the spatial fusion of the component objects detected at variable image resolutions and spatial extents. We demonstrate the efficacy of this approach in a broad area search and detection application of Surface-to-Air Missile (SAM) sites over a large search area where $>$85\% error reduction is achieved for a 100\% recall.  This new framework can be easily adapted or extended to a variety of other challenging object search and detection problems in large-scale remote sensing image datasets.

\section{STUDY AREA AND SOURCE DATA}

\begin{figure}[!t]
\centerline{\includegraphics[width=250pt]{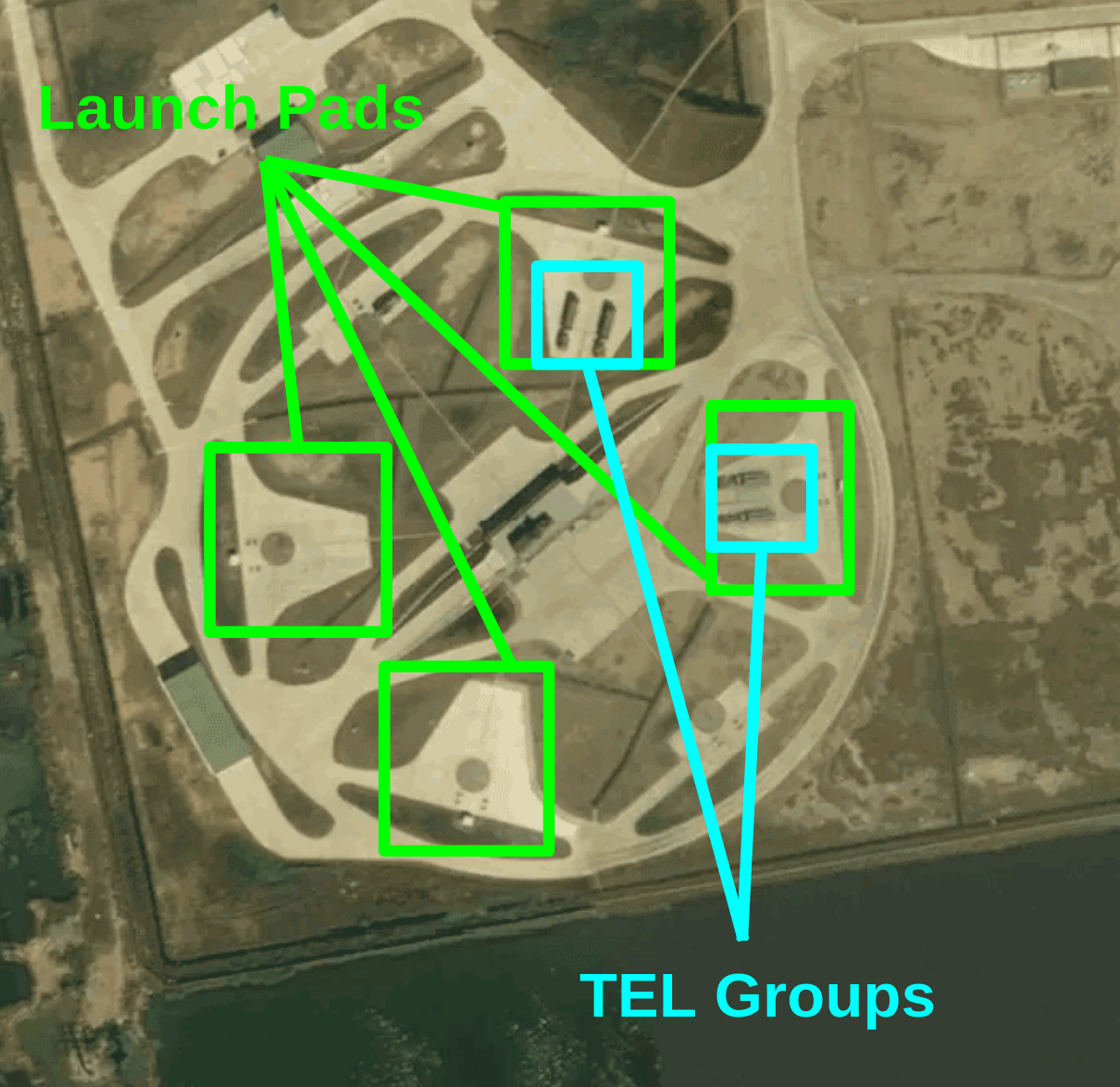}}
\caption{Example \textit{Surface-to-Air Missile (SAM) Site} with smaller-scale \textit{Launch Pad} and \textit{TEL Group} component objects.}
\label{fig:LabeledSAMSite}
\end{figure}

This study builds upon Marcum \textit{et al.}~\cite{b3} where  broad area search and detection of \textit{SAM Sites} (Fig.~\ref{fig:LabeledSAMSite}) was demonstrated over a $\sim$90,000 km\textsuperscript{2} study Area of Interest (AOI) along the SE coast of China. Key results from the prior study were:
\begin{enumerate}
\item A machine-assisted approach was used to reduce the original AOI search area by 660$X$ to only  $\sim$135 km\textsuperscript{2}.
\item The average machine-assisted search time for $\sim$2100 candidate \textit{SAM Site} locations was $\sim$42 minutes which was 81$X$ faster than a traditional human visual search.
\end{enumerate}

While Marcum \textit{et al.} used a single binary DNN detector to locate candidate \textit{SAM Sites}, here we explore the benefit of fusing multi-scale DNN detectors of smaller component objects to improve the detection of the larger encompassing \textit{SAM Site} features. 

First, a binary \textit{SAM Site} DNN detector was trained using a slightly enhanced version of the curated \textit{SAM Sites} training data in China from~\cite{b3}. To ensure blind scanning, only 101 \textit{SAM Sites} lying outside the SE China AOI were used to train the DNN. While~\cite{b3} used a 227x227 pixel chip size at 1 m GSD to train a ResNet-101 \cite{resnet} DNN, here we used a 299x299 pixel chip size at 1 m GSD for training a Neural Architecture Search Network \cite{b4} (NASNet) architecture. 

As in~\cite{b3}, negative training chip samples were selected using a 5-km offset in the four cardinal directions (i.e. N/S/E/W) for each \textit{SAM Site}. The SE China AOI has 16 known \textit{SAM Sites} which includes 2 newer \textit{SAM Sites} found in the previous study~\cite{b3}. 

\begin{figure}[ht]
     \subfloat[\textit{Empty Launch Pad}]{
       \includegraphics[width=0.1\textwidth]{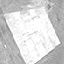}
     }
     \hfill
     \subfloat[\textit{Empty Launch Pad}]{
       \includegraphics[width=0.1\textwidth]{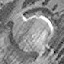}
     }
     \hfill
     \subfloat[\textit{Missile}]{
       \includegraphics[width=0.1\textwidth]{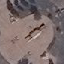}
     }
     \hfill
     \subfloat[\textit{Missile}]{
       \includegraphics[width=0.1\textwidth]{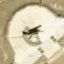}
     }
     \vfill
     \subfloat[\textit{TEL}]{
       \includegraphics[width=0.1\textwidth]{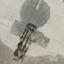}
     }
     \hfill
     \subfloat[\textit{TEL}]{
       \includegraphics[width=0.1\textwidth]{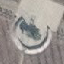}
     }
     \hfill
     \subfloat[\textit{TEL Group}]{
       \includegraphics[width=0.1\textwidth]{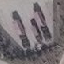}
     }
     \hfill
     \subfloat[\textit{TEL Group}]{
       \includegraphics[width=0.1\textwidth]{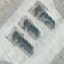}
     }
     
     \caption{Samples of \textit{SAM Site} component objects used in this study.}
     \label{fig:SampleComponents}
\end{figure}

We next developed binary DNN detectors for four different \textit{SAM Site} component objects: \textit{Launch Pads}, \textit{Missiles}, \textit{Transporter Erector Launchers} (\textit{TELs}), and \textit{TEL Groups} (two or more $\sim$co-located \textit{TELs} ) (Fig.~\ref{fig:SampleComponents}). Component binary DNN detectors were trained using curated data at 0.5 m GSD from China \textit{SAM Sites} outside the AOI. We first created negative training samples for each component using nearby image chips (similar land cover context), but outside the known spatial extent of a \textit{SAM Site}. This produced a $\sim$1:1 ratio of negative to positive component training samples (Table~\ref{table:TrainingCounts}). 

In addition, we created a second training dataset using all four components to train a combined \textit{Launch Pad} detector (empty and non-empty) knowing that the other components (e.g \textit{Missiles}, \textit{TELs} , etc.) are generally co-located with \textit{Launch Pads}. We then developed a second set of component detectors for the \textit{Missile}, \textit{TEL}, and \textit{TEL Group} object classes by combining negative training data from the other components and then randomly paring down the data to produce a $\sim$4:1 ratio of negative to positive samples (Table~\ref{table:TrainingCounts}). For the \textit{Missile} component, samples from empty \textit{Launch Pads}, \textit{TEL}, and \textit{TEL Group} and their negatives were added. However, only samples from empty \textit{Launch Pads} and \textit{Missiles} were added to the negatives for \textit{TELs}  and \textit{TEL Groups} to reduce confusion between these two components. 

Different chip sizes were used for the training samples based on known object sizes. A 128x128 pixel chip size was used for detecting both empty and combined \textit{Launch Pads} and \textit{TEL Groups}. While a 64x64 pixel chip size was used for \textit{Missiles} and \textit{TELs}. Counts for all training data are provided in Table~\ref{table:TrainingCounts} and these only include component samples outside the SE China AOI to ensure blind scanning.

\begin{table}[!t]

\caption{Summary of Curated Training Data}
\begin{center}
\begin{tabular}{|c|c|c|c|c|c|}
\hline
\rowcolor{Gray}
Object & \textit{SAM}&\textit{Launch}& \textit{Missiles}& \textit{TELs}  & \textit{TEL} \\
\rowcolor{Gray}
Class &\textit{Sites}&\textit{Pads}&&&\textit{Groups} \\
\hline
\rule{0pt}{8pt} TP & 101 & 3910\textsuperscript{1} & 1976 & 2733 & 1179 \\
\hline
\rule{0pt}{8pt}TN&404&3696&2624&2272&1054 \\
\hline
\hline
\rule{0pt}{8pt}Combo TP & n/a & 9798\textsuperscript{2} & as above & as above & as above \\
\hline
\rule{0pt}{8pt}Combo TN & n/a & 8512 & 6530 & 10,078 & 5762 \\
\hline
\multicolumn{6}{l}{
1: Empty } \\
\multicolumn{6}{l}{
2: Includes those with co-located \textit{Missiles}, \textit{TELs}, and \textit{TEL Groups}
}
\end{tabular}
\label{table:TrainingCounts}
\end{center}
\end{table}

\section{DATA PROCESSING}
\subsection{Training Data Augmentation}
Augmentation strategies from~\cite{b3} were used to train the \textit{SAM Site} DNN and all component DNNs to improve detector performance. A 144$X$ augmentation was used for all 5-fold validation experiments while a 9504$X$ augmentation was used for the final \textit{SAM Site} DNN used in the AOI scanning. To save computing time, augmentations were reduced for training the component DNNs due to the much larger sample sizes. These changes included using RGB samples only, reducing the number of rotations, using a single jitter distance, and removing the contrast augmentation. Most of the final component DNNs were trained with 648$X$ augmentations, except the combined \textit{Launch Pad} DNN used a 216$X$ augmentation. 

\subsection{Scanning and Spatial Clustering}\label{sect:scan_cluster}

We completed 5-fold cross-validation experiments for several modern DNN architectures to evaluate their performance for \textit{SAM Site} detection. The modern DNN architectures we evaluated were NASNet~\cite{b4}, Xception~\cite{xception}, ProxylessNAS~\cite{proxylessnas}, and all seven EfficientNet~\cite{efficientnet} models. \textit{SAM Site} detection results from these modern DNNs are compared to the ResNet-101 DNN results published in the Marcum et al.~\cite{b3} study (Table ~\ref{table:ArchCompare}). 

The results in Table II show that the NASNet DNN outperformed all the other DNNs for \textit{SAM Site} detection as measured by \textit{TPR}, \textit{ACC}, and \textit{AUC}. In addition to training the \textit{SAM Site} DNN detector, the NASNet DNN was used for training all component DNN detectors used throughout the rest of this study. Training for all the NASNet DNNs utilized transfer learning from ImageNet~\cite{imagenet}, Adam~\cite{adam} for optimization, and cross entropy for the objective function.

\begin{table}[!t]

\caption{Summary of \textit{SAM Site} DNN Detector Performance from 5-fold Cross-Validation. Metrics shown are True Positive Rate (\textit{TPR}), True Negative Rate (\textit{TNR}), Average Accuracy (\textit{ACC}), and Area Under the ROC Curve (\textit{AUC}).}
\begin{center}
\begin{tabular}{|c|c|c|c|c|}
\hline
\rowcolor{Gray}
\textbf{DNN} & \textit{TPR} (\%) & \textit{TNR} (\%) & \textit{ACC} (\%) & \textit{AUC} (\%) \\
\hline

\textit{ResNet-101}&94.1\%&98.8\%&96.4\%&99.4\% \\

\hline
\textit{Xception}&98.0\%&98.3\%&98.1\%&99.9\% \\
\hline

\textit{NASNet}&99.0\%&99.8\%&99.4\%&99.995\% \\
\hline

\textit{ProxylessNAS}&95.0\%&100.0\%&97.5\%&99.2\% \\
\hline

\textit{EfficientNet-B4\textsuperscript{1}}&91.1\%&99.8\%&95.4\%&99.8\% \\
\hline
\multicolumn{5}{l}{
1: Only results from the EfficientNet model with the highest \textit{AUC} are shown.
}\\

\end{tabular}
\label{table:ArchCompare}
\end{center}
\end{table}

As in~\cite{b3}, images used for broad area search for \textit{SAM Sites} in the SE China AOI were comprised of $\sim$66K 1280x1280 pixel tiles at 1 m GSD with 10\% overlap between tiles. Individual tiles were scanned by generating $\sim$19.7M image chips with 75\% overlap (25\% stride) that were then input to the NASNet models. This produced a raw detection field, $F$, of softmax outputs from the DNN. After thresholding $F$ at $\alpha=0.9$, a greatly reduced detection field, $F^\alpha$, is then used to produce an amplified spatial detection field, $\delta$. The $\delta$ is used to weight a spatial clustering of $F^\alpha$ to produce mode clusters, $F'$, within a 300-m aperture radius, $R$ (see~\cite{b3}). Cluster locations were then rank-ordered by summing the scores of all detections within a mode cluster to generate an initial set of “candidate” \textit{SAM Sites}.

A new 1280x1280 pixel tile at 0.5 m GSD centered on each candidate \textit{SAM Site}’s cluster location was then used for all component DNN scans. Component scanning outputs were also spatially clustered to generate locations and cluster scores for each component object. An aperture radius of $R$ = 32 m was used since this is approximately half the typical distance between \textit{SAM Site} launch pads in China. An alpha cut of $\alpha=0.99$ was used to generate distinct cluster locations for a given component relative to neighboring components that were present at each candidate \textit{SAM Site}. In this study we simply used \textit{a priori} knowledge for our selection of \textit{R}. However, we recognize that for other objects and/or applications the appropriate selection of \textit{R} may need to be incorporated in the technical approach, as it can be sensitive to scanning stride and target object co-location separation (e.g. vehicles parked next to each other).

Likewise, in order to determine thresholds used for the decision-theoretic approach (DTA) described in Section ~\ref{F1Sect}, a training set of 1280x1280 pixel pseudo-candidate tiles were generated that were centered about the known \textit{SAM Sites} outside the SE China AOI along with corresponding offset tile negatives.  The same scans and processes performed for candidate tiles within the SE China AOI (described above), were used for the pseudo-candidate training dataset.

\subsection{Cluster Score Normalization and Truncation}
Cluster scores from one object class to another are not necessarily comparable since they can result from objects with different physical sizes, corresponding \textit{R} values, and scanning strides.  In addition, results generated from image tile scans with different DNN input chip size and/or GSD will have a variable spatial density. Since we wish to spatially fuse, and potentially weight, the output from various component DNN detectors, the cluster scores must be normalized to bring both the \textit{SAM Site} and component detection clusters into a common reference space.  Here we use raw detection field density, i.e. the number of raw detections per unit area, as the means to achieve a common reference space prior to spatially fusing the cluster scores from the candidate \textit{SAM Sites} and their associated component detection clusters.
\subsubsection{Normalization for a Single Detection Location}
The amplified spatial detection field, $\delta$, contains an intersected volume, $\delta_n$, for each raw detection, $n$, in $F^\alpha$. $\delta_n$ is calculated as the weighted sum of scores of each $n$ with its neighboring raw detections, $p$. The weight is determined by the distance-decay function $s(p) = exp^{(-d/R)}$, where $d = haversine (p,n)< R$ and is $0$ otherwise. An approximate maximum intersection volume for a single raw detection can be calculated by integrating the truncated distance-decay function around a raw detection location. As mentioned above, $R$ and $d$ are normalized using raw detection field density. Let $R'=R/stride$ and $d'=d/stride$, where \textit{stride} is the image chip’s scanning stride distance in meters, so that $s'(p)=exp^{(-d'/R')}$. Let $s$ represent the height or the dimension of magnitude for $F$, then the approximate max intersection volume for a single raw detection can then be calculated as:

\begin{equation}
\label{eq1}
\begin{aligned}
n_{volume} & =\pi\cdot(R')^2\left(\left(\int_{1/e}^1 log^2(1/s)ds\right)+(1/e)\right)\\
& =\pi\cdot(R')^2((2-5/e)+(1/e))\\
& =\pi\cdot(R')^2(2-4/e)
\end{aligned}
\end{equation}

\subsubsection{Cluster Score Weighting/Truncation}\label{sect:MaxClusterScoreTructation}
The cluster score, $C_{score}$,  should also be limited for normalization. In the previous algorithm from~\cite{b3}, the number of raw detections in a cluster, $C$, was virtually unbounded. As a result, raw detections that were a large distance away from the cluster location can potentially contribute to $C_{score}$. We have often observed that these far away detections are commonly false positives. In order to minimize their impact, it can be beneficial to weight each raw detection within a  cluster  based  on  the  raw  detection  distance  relative  to  the cluster location. Alternately, truncation can be implemented by assigning each raw detection within a haversine distance \textit{R} a weight of 1 and outside of \textit{R} a weight of 0. The $C_{score}$ is then a weighted sum of the raw detection inference scores with their respective weights (Procedure 1). In Section~\ref{sect:penalty}, we discuss the possibility of applying a negative penalty weight to all raw detections with a haversine distance greater than $R$. 

\floatname{algorithm}{Procedure}

\begin{algorithm}
\caption{Object Detection Cluster Ranking with Normalized Scores and Optional Penalty} 
\begin{algorithmic}\label{Procedure1}
\State\hspace{-1.25em} \textbf{Input:} Alpha-cut $F^\alpha$, Mode-Cluster $F'$ 
\State\hspace{-1.25em} \textbf{Output:} Ranked Clusters $C_i$, where $C_i<C_{i+1}$ 
\State\hspace{-1.25em} \textbf{begin}
    \State $i:=0$
    \While {$F'=\emptyset$}
    \State $p:=pop( F')$
    \State $C_i:=p$ // Init. $C_i$ with chip, $p$
    \State $N^\alpha:=NN(p,F^\alpha,R)$
    \State $N:=NN(p,F',R)$
    \ForAll{$n \in N(p)$}
        \State {$C_i:=\{C_i,n\}$ $F'.remove(n)$}
        \If {$n \subset N^\alpha(p)$} 
            \State {$n_{weight}=1$}
        \Else
            \State $n_{weight}=$ penalty // 0 if no penalty
        \EndIf
    \EndFor
    \State $C_i.score =(C_{norm})^{-1}\cdot\sum_{n \in C_i} n_{weight}\cdot\delta_n$
    \State $i$++
    \EndWhile
    \State $\{C_i\}:=$\textbf{sort} $C_{\forall i}$ by score, descending
\State\hspace{-1.25em} \textbf{end}
\end{algorithmic}
\end{algorithm}

\subsubsection{Approximate Max Cluster Score}
The number of raw detections within $R$ can be seen as a Gauss Circle Problem. Thus, the max number of raw detections within the aperture area surrounding a cluster location can be approximated in terms of raw detection field density as: 
\begin{equation}
n_{max\_p}=\pi\cdot(R')^2  
\label{eq2}
\end{equation}
Using equations (\ref{eq1}) and (\ref{eq2}), a normalizing cluster factor can be calculated as:
\begin{equation}
\begin{aligned}
                 C_{norm} &=n_{volume}\cdot n_{max\_p} \\  
                            &=\pi^2\cdot(R')^4(2-4/e)
\label{eq3}
\end{aligned}
\end{equation}

\subsection{Over-Detection Penalty}\label{sect:penalty}
\begin{figure}[!t]
\centerline{\includegraphics[width=250pt]{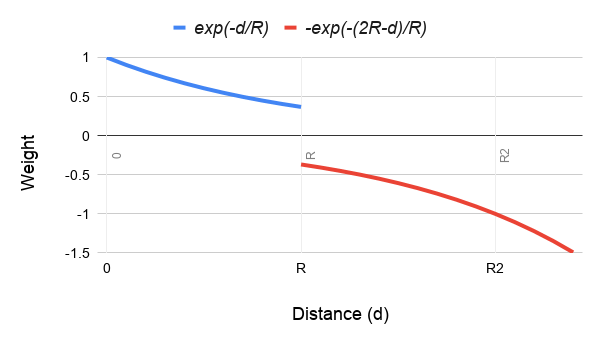}}
\caption{Distance-decay functions used for calculating local clustering scores.  The function $exp(-d/R)$ (blue) is used as a weight when summing raw detections within distance $R$. The function $-exp(-(2R-d)/R)$ (red) is used to calculate an exponential penalty weight for raw detections outside $R$.}
\label{fig:DistanceDecay}
\end{figure}

In previous work we have observed FP hotspots, i.e. large numbers of spatially co-occurring false positive detections. In order to mitigate this potential problem, a penalty can be applied when computing $C_{score}$. As mentioned in Section \ref{sect:MaxClusterScoreTructation}, instead of using a weight of 0 when $d > R$, a negative weight can be applied. We explored two types of penalty assignments. The first used a flat weight of -1. The second is similar to the distance-decay function, however the sign was changed to negative and increases in value exponentially as $d$ increases (Fig.~\ref{fig:DistanceDecay}). The penalty is calculated using the following formula:  $s(p) = -exp(-(2R-d)/R)$.

\begin{figure}[!t]

\centerline{\includegraphics[width=250pt]{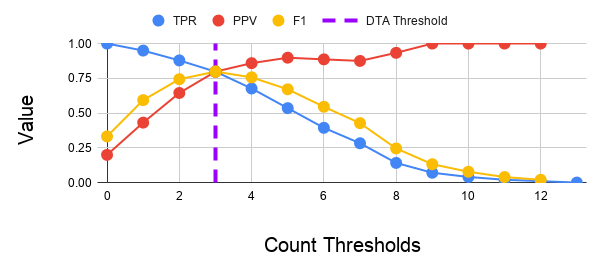}}
\caption{True Positive Rate (\textit{TPR} or recall), Positive Predictive Value (\textit{PPV} or precision), and \textit{F1} score versus the threshold for the cluster count of \textit{TEL} cluster centers within 150 m of a candidate \textit{SAM Site} location.  In this example, the value 3 was  used for the final threshold as shown in Table~\ref{table:SampleDTAThresholds}.}
\label{fig:F1score}
\end{figure}

\begin{figure*}[!]
\includegraphics[width=\textwidth]{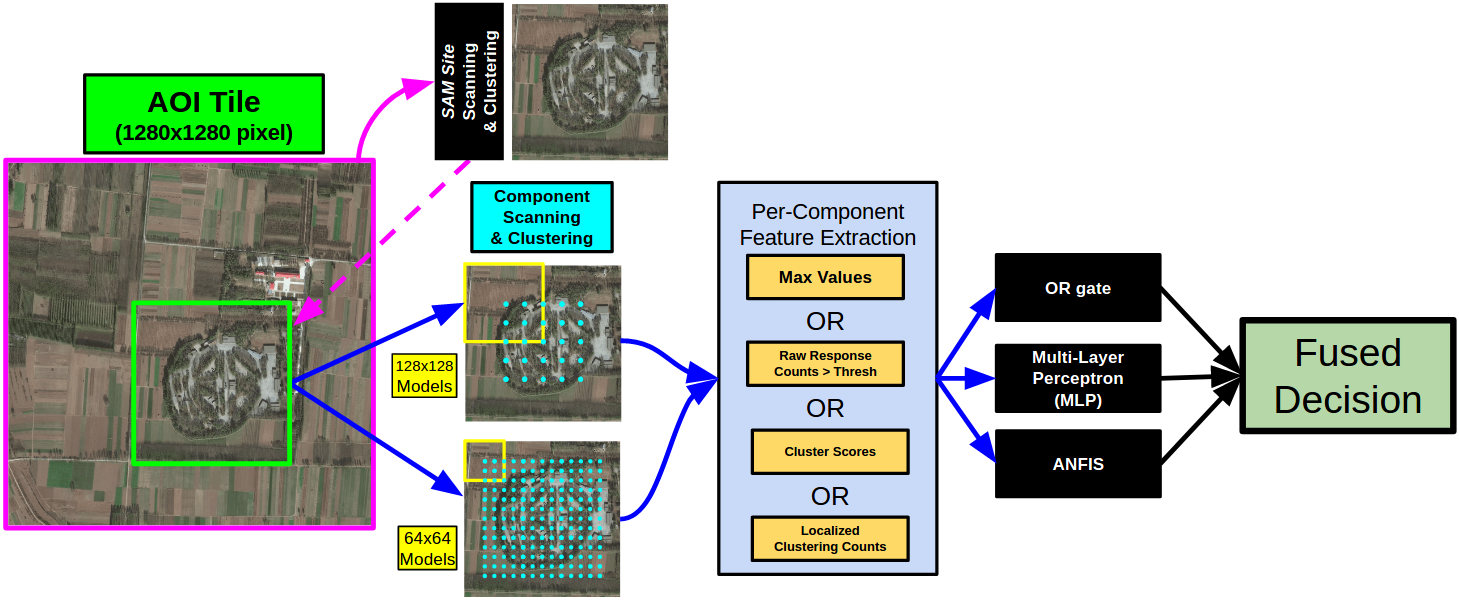}
\caption{Processing flow chart for decision-level fusion of multiple component object detections.}
\label{fig:ScoreFusionFlow}
\end{figure*}

\subsection{Decision-Theoretic Approach for Optimization}\label{F1Sect}
In order to make discrete decisions, we used the DTA~\cite{b6} advocated by Lewis~\cite{b7} that computes thresholds based on the optimal prediction of a model to obtain the highest expected \textit{F}-measure.  In this study, decision thresholds were selected based on the optimization of the \textit{F1} score from features extracted from the pseudo-candidate training dataset.
(Fig.~\ref{fig:F1score}). Optimal \textit{F1} score thresholds were determined through empirical analysis and selected examples are provided (Table ~\ref{table:SampleDTAThresholds}).
\begin{table}[!t]

\caption{Sample thresholds calculated by DTA.}
\begin{center}
\begin{tabular}{|c|c|c|c|c|c|}
\hline
\rowcolor{Gray}
\textit{Feature}&\textit{Empty}&\textit{Combo}&\textit{Missiles}&\textit{TELs}&\textit{TEL}\\
\rowcolor{Gray}
\textit{Type}&\textit{LPs}&\textit{LPs}&&&\textit{Groups}\\\hline
Cluster Count&2&1&1&3&1\\\hline
Raw Count&5&4&4&15&1\\\hline
Raw Max&1.00000&0.99954&1.00000&1.00000&0.58236\\\hline
\multicolumn{6}{|c|}{Including Component Negatives}\\\hline
Cluster Count&n/a&n/a&1&1&1\\\hline
Raw Count&n/a&n/a&2&2&1\\\hline
Raw Max&n/a&n/a&0.99989&0.98450&0.60315\\\hline
\end{tabular}
\label{table:SampleDTAThresholds}
\end{center}
\end{table}

\section{FIVE-FOLD EXPERIMENT RESULTS}
Five-fold cross validation experiments were performed for the training datasets in Table ~\ref{table:TrainingCounts}. The results provided in Table~\ref{table:FiveFoldSAM} show an average \textit{F1} score $> 99\%$ for the baseline dataset and $> 98\%$ for the dataset with component negatives. The decrease in \textit{F1} score for the DNNs with component negatives was anticipated given the inclusion of objects in the negative training data that were visually similar to the component object that a given DNN was trained to detect.

\begin{table}[!t]

\caption{NASNet five-fold cross validation results for DNN models of \textit{SAM Sites} and each component, including component models with negative component data. Metrics shown are True Positive Rate (\textit{TPR}), True Negative Rate (\textit{TNR}), \textit{F1} score, and Standard Deviation (\textit{SD}).}
\begin{center}
\begin{tabular}{|c|c|c|c|c|}
\hline
\rowcolor{Gray}
\textit{\textbf{Object Class}}&\textit{TPR} (\%) &\textit{TNR} (\%)&\textit{F1} (\%)&\textit{SD} \\\hline
\textit{SAM Site}&99.00&99.75&99.39&1.06\\\hline
\textit{Empty LPs}&99.80&99.70&99.65&0.2\\\hline
\textit{Combo LPs}&99.74&99.74&99.74&0.15\\\hline
\textit{Missiles}&99.8&99.46&99.63&0.24\\\hline
\textit{TELs} &99.72&99.38&99.55&0.32\\\hline
\textit{TEL Groups}&99.41&99.43&99.42&0.21\\\hline
\multicolumn{5}{|c|}{Including Component Negatives}\\\hline
\textit{Missiles}&97.42&99.66&98.52&0.72\\\hline
\textit{TELs} &97.51&99.37&98.42&0.5\\\hline
\textit{TEL Groups}&96.78&99.6&98.15&1.1\\\hline
\end{tabular}
\label{table:FiveFoldSAM}
\end{center}
\end{table}

NASNet significantly outperformed ResNet-101 for scanning the SE China AOI for \textit{SAM Sites} (Table~\ref{table:SAMCand}). This is consistent with the cross-validation results given in Table~\ref{table:FiveFoldSAM}. NASNet had $\sim$44$X$ fewer \textit{SAM Site} candidate locations after the 0.9 alpha-cut (Section ~\ref{sect:scan_cluster}). Further, while both DNNs correctly located all 16 known \textit{SAM Sites} (e.g. TPs) in the SE China AOI, NASNet had 6$X$ fewer candidates compared to ResNet-101 while the average TP cluster rank (Table~\ref{table:SAMCand}) was also $\sim$3$X$ lower. 
\begin{table}[!t]

\caption{Spatial clustering results from DNN scanning of the SE China AOI for candidate \textit{SAM Sites}. Given values are pre-cluster counts over $\alpha$-cut threshold ($F^\alpha$), post-cluster counts, and average True Positive (\textit{TP}) cluster rank.}
\begin{center}
\begin{tabular}{|c|c|c|c|c|c|c|}
\hline
\rowcolor{Gray}
DNN Architecture&$F^\alpha$&C&AVG \textit{TP}\\
\rowcolor{Gray}
\& Post-Procsessing&Count&Count&Cluster Rank\\\hline
ResNet-101~\cite{b3}&~93,000&~2100&181.9\\\hline
NASNet&2079&354&62.8\\\hline
NASNet w/ norm&2079&354&62.8\\\hline
NASNet w/ norm and penalty&2079&354&62.8\\\hline
\end{tabular}
\label{table:SAMCand}
\end{center}
\end{table}
\section{DECISION-LEVEL COMPONENT METRIC FUSION}\label{sect:compFusion}
This section describes the feature selection and fusion techniques used to reduce the number of candidate \textit{SAM Sites} that could then be presented for human review in machine-assisted analytic workflows. An overview of the processing flow is provided in Fig.~\ref{fig:ScoreFusionFlow}. 

\subsection{Component Features}
Five different feature types were used in~\cite{b5} for decision-level fusion of component objects for improving the final detection of construction sites. Here we tested feature types that used the \textit{F1} score optimization from~\cite{b5} and represent the first three feature types listed below. We used the normalized cluster scores from the spatial clustering as an additional feature type. To maintain consistency between techniques employed in this study, only inference responses within a 150 m radius of the candidate \textit{SAM Site} location were used. The feature types that were evaluated were:

\begin{enumerate}
\item Maximum raw inference detection response (confidence value) for each component.
\item Count of raw inference detections for each component retained within the reduced field ($F^\alpha$).  
\item Count of clusters produced  for each component.
\item Sum of normalized cluster scores for each component.
\end{enumerate}

\subsection{Decision-Level Fusion Techniques}
Baseline results for the candidate \textit{SAM Site} locations are first computed using only the spatial cluster outputs of the NASNet \textit{SAM Site} detector. We then tested how each individual component would perform using the various feature types. \textit{SAM Site} cluster scores were excluded because the pseudo-candidate training dataset was NOT generated through scanning and clustering. Consequently, some of the pseudo-candidates would have no cluster within a sufficient radius of the \textit{SAM Site} center location.

Three data fusion techniques were tested:
\begin{enumerate}
\item \textbf{Decision Tree:} A simple decision tree (see~\cite{b5}) was used to combine the decisions generated for each component using DTA. However, unlike~\cite{b5}, this study does not use an alpha-cut threshold since this was part of the spatial clustering algorithm. Therefore, the decision tree is simplified to a digital logic OR gate with the DTA decisions as binary inputs.
\item \textbf{Multi-Layer Perceptron (MLP):}  A feature vector was created for each candidate \textit{SAM Site} location and used as input for training and validation. The MLP architecture consisted of two fully connected hidden layers of 100 nodes. We also tested normalization and feature bounds before being used as input based on the thresholds from DTA optimization  (Section~\ref{F1Sect}).
\item \textbf{ANFIS:} A first order Takagi-Sugeno-Kang (TSK) adaptive neuro-fuzzy inference system (ANFIS)~\cite{b24}~\cite{b22}~\cite{b23} was utilized. The goal is to explore a neural encoding and subsequent optimization of expert knowledge input. Specifically, five IF-THEN rules were used whose IF components (aka rule firing strengths) were derived from the expert knowledge from the Decision Tree in 1) above. The consequent (i.e., ELSE) parameters of ANFIS were optimized via backpropagation~\cite{b24}. The reader can refer to~\cite{b25}~\cite{b27} and~\cite{b28} for an in-depth discussion of the mathematics, optimization, and robust possibilistic clustering-based initialization of ANFIS. Finally, the output decision threshold was chosen through DTA. 

\end{enumerate}

\begin{figure*}[!t]
\includegraphics[width=\textwidth]{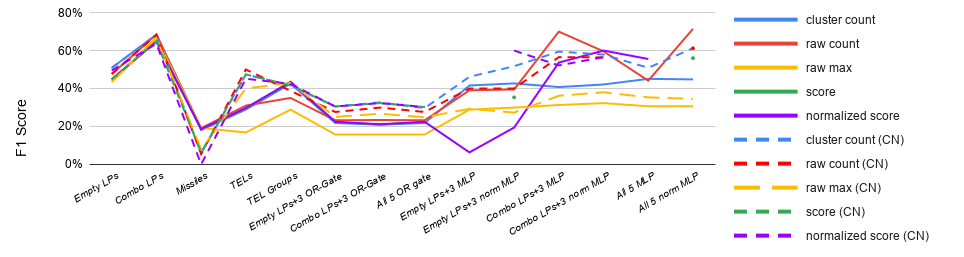}
\caption{Comparison of \textit{F1} scores produced for candidate \textit{SAM Site} locations from different fusion techniques. Techniques include individual component threshold from DTA as well as component fusion using an OR gate and MLP. Note that “(CN)“ at the end of the feature type label in the key indicates that component negative models were used in the processing. Gaps in scores occur when the MLP was unable to train on a given feature type.}
\label{fig:F1ScoreComp}
\end{figure*}
\begin{figure*}[!t]
\centering
\includegraphics[width=\textwidth]{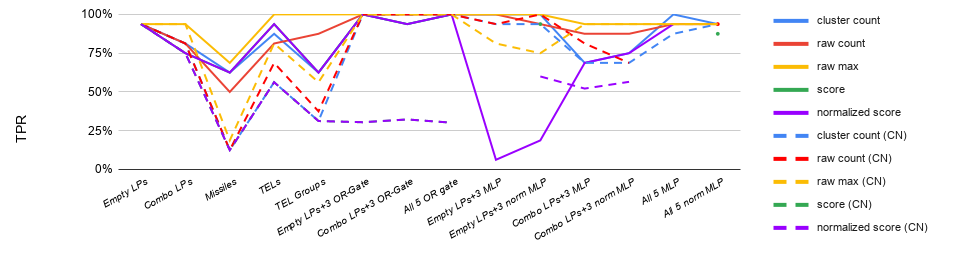}
\caption{Comparison of True Positive Rate (\textit{TPR}) produced for candidate \textit{SAM Site} features from different fusion techniques.}
\label{fig:TPRCompare}
\end{figure*}

The different \textit{Launch Pad} detector types were tested independently and in combination during the fusion step with the other three component types (i.e. \textit{Missiles}, \textit{TELs}, and \textit{TEL Groups}):

\begin{itemize}
    \item Empty \textit{Launch Pads} plus three (\textit{Empty LPs+3})
    \item Combined \textit{Launch Pads} plus three (\textit{Combo LPs+3})
    \item \textit{Empty LPs} and \textit{Combo LPs} plus three (All 5)
\end{itemize}

\subsection{MLP Input Data Normalization}
We found that the MLPs had some difficulty training with datasets that had larger values, so we used the common practice of linearly scaling and bounding to constrain the data to fall within the range $[-1,1]$. Let $v_i$ be the vector of values over the entire dataset for component $i$ for a given feature and let $t_i$ be the DTA thresholds computed for component $i$, then the normalized and bounded vector $v_i'$ can be defined as follows:
 \begin{equation}
    v_i'=
    \begin{cases}       
         \textrm{if }&(v_i-t_i)/t_i  \in [-1,1], \textrm{ then } (v_i-t_i)/t_i \\
        \textrm{if }&(v_i-t_i)/t_i <-1, \textrm{ then }-1 \\     
        \textrm{if }&(v_i-t_i)/t_i >1   , \textrm{ then }1 
   
    \end{cases}       
 \end{equation}

\subsection{Results \& Observations}

Over 200 different combinations of data feature types, component combinations, and fusion techniques were tested in this study to improve the final detection of candidate \textit{SAM Sites}.

Evaluation of the \textit{F1} score improvements (Table~\ref{table:TopF1Score}) shows that decision-level component fusion can reduce the relative error rate by up to 96.75\%. It was somewhat surprising that the Raw Count feature  generated five out of the top six best results. Although \textit{Combo LPs} were only able to generate an \textit{F1} score of 68.4\% using DTA, the neural approaches (MLP and ANFIS) were able to do slightly better using multiple components where the top results fused all 5 components in an MLP to yield an \textit{F1} score of 71.4\%. Comparisons of \textit{F1} scores for different feature types and fusion techniques can be found in Fig.~\ref{fig:F1ScoreComp}.

However, when performing a broad area search for a very rare object (low geographic occurrence rate), it is often desirable to sacrifice some error reduction in order to achieve a higher \textit{TPR}. The results in Table~\ref{table:TopTPR} show that the highest \textit{F1} score is 45.1\% while achieving a \textit{TPR} of 100\%. Although this \textit{F1} score is less than half of the maximum in Table ~\ref{table:TopF1Score}, this technique still achieved a 88.5\% relative error reduction compared to the baseline (no component fusion) results for the candidate \textit{SAM Site} locations within the SE China AOI. These scores were produced using Cluster Count features and the \textit{All 5} component combination as inputs to a simple MLP. It is also worth noting that four of the top five scores used the \textit{Empty LPs+3} component combination. Comparisons of TPRs for different feature types and fusion techniques can be found in Fig.~\ref{fig:TPRCompare}.

\begin{figure}[t]
\setlength{\belowcaptionskip}{-10pt}
\centerline{\includegraphics[width=250pt]{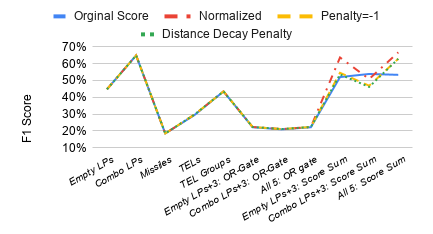}}
\caption{\textit{F1} score results for DTA thresholds of original cluster scores, normalized cluster scores, cluster scores with a penalty of -1 and distance-decay penalty with $R=150$ m.}
\label{fig:PenCompare}
\end{figure}

It was also observed that cluster score truncation and normalization was able to improve the \textit{F1} scores for DTA when fusing multiple component detectors. However, the introduction of negative score penalty did not improve the score further (Fig ~\ref{fig:PenCompare}), while introducing expert weighting (described in Section~\ref{sect:Ranking}) also showed no improvement for the \textit{F1} scores. 

\begin{table*}[!t]

\caption{
Experiment results with highest \textit{F1} Scores. The first line after the header (in red) are the results for \textit{SAM Site} detection \underline{without} error reduction from spatial fusion of any component feature type(s). The highest \textit{F1} scores were achieved by fusing multiple components using neural learning techniques (MLP or ANFIS). Also, raw detection counts (pre-clustering) showed the most separability. All top solutions achieved a relative error reduction of greater than $96\%$. These results would be optimal if error reduction was the primary goal. The error rate includes both false positives and false negatives.
}
\begin{center}
\begin{tabular}{|c|c|c|c|c|c|c|c|c|c|c|}
\hline
\rowcolor{Gray}
Components&Feature&Processing&Component&\textit{TP}&\textit{FP}&\textit{TPR}&\textit{PPV}&\textit{F1} score&Error / km\textsuperscript{2}&Relative Error\\
\rowcolor{Gray}
&Type&Technique&Negatives&&&(Recall)&(Precision)&&(x$10^{-3}$)&Reduction\\\hline
\textcolor{red}{\textit{SAM Sites}}&\multicolumn{3}{|c|}{\textcolor{red}{BASELINE-NO COMPONENTS}}&\textcolor{red}{16}&\textcolor{red}{338}&\textcolor{red}{100.00\%}&\textcolor{red}{4.52\%}&\textcolor{red}{8.65\%}&\textcolor{red}{3.080}&\textcolor{red}{n/a}\\\hline
All 5&Raw Counts&MLP&NO&15&11&93.75\%&57.69\%&71.43\%&0.109&96.45\%\\\hline
\textit{Combo LPs+3}&Raw Counts&ANFIS&NO&13&8&81.25\%&61.90\%&70.27\%&0.100&96.75\%\\\hline
\textit{Combo LPs+3}&Raw Counts&MLP&NO&14&10&87.50\%&58.33\%&70.00\%&0.109&96.45\%\\\hline
\textit{Combo LPs}&Cluster Count&DTA&n/a&13&9&81.25\%&59.09\%&68.42\%&0.109&96.45\%\\\hline
\textit{Combo LPs}&Raw Count&DTA&n/a&13&9&81.25\%&59.09\%&68.42\%&0.109&96.45\%\\\hline
All 5&Raw Count&ANFIS&NO&13&9&81.25\%&59.09\%&68.42\%&0.109&96.45\%\\\hline
\end{tabular}
\label{table:TopF1Score}
\end{center}
\end{table*}
\begin{table*}[!t]
\caption{Experiment results with highest \textit{F1} scores while maintaining a \textit{TPR} of 100\%. The highest \textit{F1} scores resulted from fusing a feature from all components with a simple MLP. Also, Cluster Count features yielded the top results. All top solutions show a reduction of relative error between $85.2\%-88.5\%$ which is 3$X$ the error rate shown in Table~\ref{table:TopF1Score}.}
\begin{center}
\begin{tabular}{|c|c|c|c|c|c|c|c|c|c|c|}
\hline
\rowcolor{Gray}
Components&Feature&Processing&Component&\textit{TP}&\textit{FP}&\textit{TPR}&\textit{PPV}&\textit{F1} score&Error/ km\textsuperscript{2}&Relative Error\\
\rowcolor{Gray}
&Type&Technique&Negatives&&&(Recall)&(Precision)&&(x$10^{-3}$)&Reduction\\\hline
\textcolor{red}{\textit{SAM Sites}}&\multicolumn{3}{|c|}{\textcolor{red}{BASELINE-NO COMPONENTS}}&\textcolor{red}{16}&\textcolor{red}{338}&\textcolor{red}{100\%}&\textcolor{red}{4.52\%}&\textcolor{red}{8.65\%}&\textcolor{red}{3.080}&\textcolor{red}{n/a}\\\hline
All 5&Cluster Count&MLP&NO&16&39&100\%&29.09\%&45.07\%&0.355&88.46\%\\\hline
\textit{Empty LPs+3}&Cluster Count&MLP (Normalized)&NO&16&43&100\%&27.12\%&42.67\%&0.392&87.28\%\\\hline
\textit{Empty LPs+3}&Cluster Count&MLP&NO&16&45&100\%&26.23\%&41.56\%&0.410&86.69\%\\\hline
\textit{Empty LPs+3}&Raw Count&MLP (Normalized)&YES&16&48&100\%&25.00\%&40.00\%&0.437&85.80\%\\\hline
\textit{Empty LPs+3}&Raw Count&MLP&NO&16&50&100\%&24.24\%&39.02\%&0.456&85.21\%\\\hline
\end{tabular}
\label{table:TopTPR}
\end{center}
\end{table*}

Additionally, in general there was improvement in \textit{F1} scores for models trained with component negatives, however these improvements came at a sacrifice in \textit{TPR} and only have one appearance in the Tables~\ref{table:TopF1Score} and~\ref{table:TopTPR}.  This can be interpreted as ambiguity being introduced to the dataset by essentially asking the detector to ignore the background (i.e. the \textit{Launch Pad}) and focus on the smaller component.

\section{COMPONENT METRICS FUSION FOR IMPROVING CANDIDATE \textit{SAM SITE} RANKINGS}
This section discusses techniques, observations, and results used to re-rank candidate \textit{SAM Sites} for utilization in machine-assisted human analytic workflows. The objective is to utilize the component detection clusters to re-rank the candidate \textit{SAM Sites} such that true \textit{SAM Sites} appear higher in a rank-ordered list relative to a baseline ranking derived only from the candidate \textit{SAM Sites’} cluster scores (Table~\ref{table:RANKBaseline}). An overview of the processing flow is given in Fig.~\ref{fig:RankFlow}.

\subsection{Candidate Site and Component Score Spatial Fusion} \label{sect:Ranking} 
\begin{figure}[!t]
\centerline{\includegraphics[width=250pt]{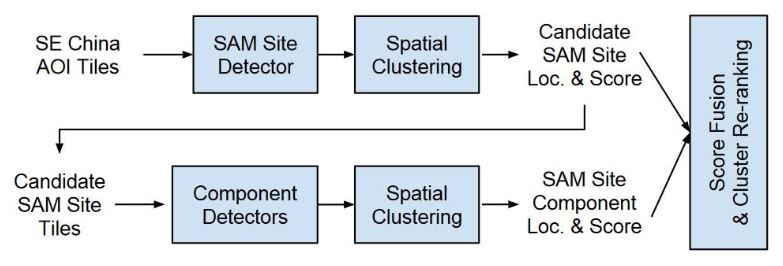}}
\caption{Process flow used for improved ranking of candidate \textit{SAM Sites}.}
\label{fig:RankFlow}
\end{figure}

Normalized cluster scores for candidate \textit{SAM Sites} and all components found within \textit{R} are summed using uniform or human expert provided weights (Fig.~\ref{fig:RankFlow}). Expert weights were only used when fusing all four components with its corresponding candidate \textit{SAM Site}. The weights were: 4 for \textit{Launch Pads}, 2 for \textit{TEL Groups}, and 1 for \textit{Missiles}, \textit{TELs}, and \textit{SAM Sites}.

\subsection{Results \& Observations}

The \textit{TEL} detector rendered the most improvement in the average cluster rank of known \textit{SAM Sites} (\textit{TPs}) compared to fusion with any other single component detector (Table~\ref{table:RANKBaseline}). This, coupled with the \textit{Combo LPs} detector and other component detectors trained with expert weighting (Section~\ref{sect:Ranking}) improved the average cluster rank of known \textit{SAM Sites} (\textit{TPs}) to 15.9 (Table~\ref{table:RANK}). This is $\sim$4$X$ better than the average rank for \textit{SAM Sites} without spatial fusion of the component object cluster scores.

We observed that the addition of normalization and penalty had no detectable impact on the known \textit{SAM Site} TP average cluster rank. This indicates minimal FP presence and/or uniformly distributed FP noise within the candidate \textit{SAM Site} locations generated by the spatial clustering algorithm.

Component negative models improved the ranking results compared to the \textit{SAM Site} score alone, but not as well as models trained without component negatives. Again, this can be interpreted as ambiguity being introduced to the dataset by essentially asking the detector to ignore the background (i.e. the \textit{Launch Pad}) and focus on the smaller component. 

\begin{table}[!t]

\caption{Average rank of known \textit{SAM Sites} (TPs) in SE China AOI from fusing cluster scores from a single component object class with a baseline candidate \textit{SAM Site} cluster score.}
\begin{center}
\begin{tabular}{|c|>{\centering\arraybackslash}m{1cm}|c|c|c|c|c|}
\hline
\rowcolor{Gray}
&&\multicolumn{5}{|c|}{ with Single Component Fusion}\\\cline{3-7}
\rowcolor{Gray}
\rule{0pt}{25pt}&\textit{SAM Site} Only&\rotatebox[origin=c]{90}{\textit{Empty LPs}}&\rotatebox[origin=c]{90}{\textit{Combo LPs}}&\rotatebox[origin=c]{90}{\textit{Missiles}}&\rotatebox[origin=c]{90}{\textit{TELs} }&\rotatebox[origin=c]{90}{\textit{TEL Groups}}\\\hline
ResNet-101\cite{b3}&139.9&n/a&n/a&n/a&n/a&n/a\\\hline
NASNet&62.8&36.4&40.8&43.0&28.0&46.1\\\hline
w/ Norm&62.8&34.3&34.4&43.6&28.1&47.3\\\hline
w/ Norm \& Penalty&62.8&34.0&34.3&43.6&\cellcolor{yellow}\textbf{27.9}&47.1\\\hline
\multicolumn{7}{|c|}{Including  Component Negatives}\\\hline
w/ Norm&n/a&n/a&n/a&79.1&28.8&51.6\\\hline
w/ Norm \& Penalty&n/a&n/a&n/a&79.1&28.7&51.48\\\hline
\end{tabular}
\label{table:RANKBaseline}
\end{center}
\end{table}

\begin{table}[!t]

\caption{Average rank of known \textit{SAM Sites} (TPs) in SE China AOI from fusing cluster scores from all four component object classes with the baseline candidate \textit{SAM Site} cluster score.}
\begin{center}
\begin{tabular}{|c|c|c|c|c|}
\hline
\rowcolor{Gray}

&\multicolumn{2}{|c|}{\textit{Empty LPs}}&\multicolumn{2}{|c|}{\textit{Combo LPs}}\\\cline{2-5}
\cellcolor{Gray}\rule{0pt}{30pt}&\rotatebox[origin=c]{90}{\cellcolor{Gray}\shortstack{Unweighted \\ Fusion}}&
\rotatebox[origin=c]{90}{{\cellcolor{Gray}\shortstack{Weighted \\ Fusion}}}\rotatebox[origin=c]{90}&
\rotatebox[origin=c]{90}{\cellcolor{Gray}\shortstack{Unweighted \\ Fusion}}&
\rotatebox[origin=c]{90}{\cellcolor{Gray}\shortstack{Weighted \\ Fusion}}\\\hline
NASNet&{26.3}&{21.4}&{25.3}&{22.9}\\\hline
w/ Norm&{20.3}&{22.9}&{17.9}&{\cellcolor{yellow}\textbf{15.9}}\\\hline
w/ Norm \& Penalty&{19.9}&{22.5}&{17.8}&{16.0}\\\hline
\multicolumn{5}{|c|}{Including  Component Negatives}\\\hline
w/ Norm&{24.8}&{24.9}&{18.1}&{16.8}\\\hline
w/ Norm \& Penalty&{24.1}&{24.9}&{18.1}&{16.8}\\\hline
\end{tabular}
\label{table:RANK}
\end{center}
\end{table}

\section{CONCLUSION AND FUTURE WORK}
This study extended the work in~\cite{b3} where a combination of a DNN scanning and spatial clustering was used to perform a machine-assisted broad area search and detection of \textit{SAM Sites} in a SE China AOI of $\sim$90,000 km\textsuperscript{2}. 

Here we significantly improved upon this prior study by using multiple DNNs to detect smaller component objects, e.g. \textit{Launch Pads}, \textit{TELs}, etc. belonging to the larger and more complex \textit{SAM Site} feature. Scores computed from an enhanced spatial clustering algorithm were normalized to a reference space so that they were independent of image resolution and DNN input chip size. A variety of techniques were then explored to fuse the DNN detections from the multiple component objects to improve the final detection and retrieval (ranking) of DNN detections of candidate \textit{SAM Sites}. Key results from this effort include: 
\begin{enumerate}
\item Spatial fusion of DNN detections from multiple component objects using neural learning techniques that maximize the \textit{F1} score reduced an initial set of $\sim$350 \textit{SAM Site} detections (Table~\ref{table:SAMCand}) to only $\sim$25 candidate \textit{SAM Sites} (Table~\ref{table:TopF1Score}).
\item  An alternate spatial fusion approach from that used in 1) reduced the overall error rate by $>$85\% while preserving a 100\% \textit{TPR} (Table~\ref{table:TopTPR}) and also reduced the initial set of detections to $\sim$55-60 candidate \textit{SAM Sites}.
\item  The average rank of 16 known \textit{SAM Sites} (\textit{TP}s) in a list of $\sim$350 candidate \textit{SAM Sites} was improved by $\sim$9$X$ (Tables~\ref{table:RANKBaseline} and~\ref{table:RANK}) compared to the previous study~\cite{b3}. 
\end{enumerate}

In future work we plan to A) apply this approach to a variety of other challenging object search and detection problems in large-scale remote sensing image datasets, B) investigate data-driven optimization of the component fusion weights and compare performance vs. human-expert provided weights, C) extend this approach to include fusion of multi-temporal DNN detections, D) extend this approach to include fusion of multi-source DNN detectors applied to high-resolution EO/MS and SAR imagery, and E) explore how to use more sophisticated fusion techniques (similar to ANFIS) to maintain \textit{TPR} while achieving even higher error reduction.


\begin{thebibliography}{00}
\bibitem{b1} G. J. Scott, K. C. Hagan, R. A. Marcum, J. A. Hurt, D. T. Anderson, and C. H. Davis, “Enhanced fusion of deep neural networks for classification of benchmark high-resolution image datasets,’’ \textit{IEEE Geoscience \& Remote Sensing Letters}, Vol. 15, No. 9, pp. 1451-1455, 2018, DOI: 10.1109/LGRS.2018.2839092. 
\bibitem{b2} J. A. Hurt, G. J. Scott, D. T. Anderson, C. H. Davis, “Benchmark meta-dataset of high-resolution remote sensing imagery for training robust deep learning models in machine-assisted visual analytics,’’ \textit{2018 IEEE Applied Imagery Pattern Recognition Workshop (AIPR)}, 9-11 October, 2018.
\bibitem{b8} G. J. Scott, M. R. England, W. A. Starms, R. A. Marcum, and C.H. Davis (2017), “Training deep convolutional neural networks for land cover classification of high-resolution imagery,’’ \textit{IEEE Geoscience \& Remote Sensing Letters}, Vol. 14, No. 4, pp. 549-553, DOI: 10.1109/LGRS.2017. 2657778.
\bibitem{b9} G. J. Scott , R. A. Marcum, C. H. Davis and T. W. Nivin, “Fusion of deep convolutional neural networks for land cover classification of high-resolution imagery,’’ \textit{IEEE Geoscience \& Remote Sensing Letters}, Vol. 14, No. 9, 2017, pp. 1638-1642, DOI: 10.1109/LGRS. 2017.2722988.
\bibitem{b10} Y. Yang and S. Newsam, “Bag-of-visual words and spatial extensions for land-use classification,’’ \textit{Proc. ACM SIGSPATIAL Int. Conf. Adv. Geogr. Inf. Syst.}, 2010, pp. 270–279.
\bibitem{b11} G. Sheng, W. Yang, T. Xu, and H. Sun, “High-resolution satellite scene classification using a sparse coding based multiple feature combination,’’ \textit{International Journal of Remote Sensing}, Vol. 33, No. 8, 2012, pp. 2395–2412.
\bibitem{b13} G. Cheng, J. Han, and X. Lu., “Remote sensing image scene classification: benchmark and state of the art,’’ \textit{Proceedings of the IEEE}, Vol. 105, No. 10, 2017, pp. 1865-1883, DOI: 10.1109/JPROC.2017.2675998.
\bibitem{yolo} J. Redmon, S. Divvala, R. Girshick and A. Farhadi, “You Only Look Once: Unified, Real-Time Object Detection,’’ \textit{2016 IEEE Conference on Computer Vision and Pattern Recognition (CVPR)}, Las Vegas, NV, 2016, pp. 779-788, doi: 10.1109/CVPR.2016.91.
\bibitem{rcnn} R. Girshick, J. Donahue, T. Darrell, J. Malik, “Rich Feature Hierarchies for Accurate Object Detection and Semantic Segmentation,’’ \textit{2014 IEEE Conference on Computer Vision and Pattern Recognition}, 2014, pp. 580–587.
\bibitem{fastrcnn} R. Girshick, “Fast R-CNN,’’ \textit{The 2015 IEEE International Conference on Computer Vision (ICCV)}, 2015, pp. 1440–1448,  DOI: 10.1109/ICCV.2015.169 
\bibitem{fasterrcnn} S. Ren, K. He, R. Girshick and J. Sun, “Faster R-CNN: Towards Real-Time Object Detection with Region Proposal Networks,’’ \textit{IEEE Transactions on Pattern Analysis and Machine Intelligence}, vol. 39, no. 6, 2017, pp. 1137-1149, DOI: 10.1109/TPAMI.2016.2577031.
\bibitem{yolov3} J. Redmon, A. Farhadi, “YOLOv3: An Incremental Improvement,’’ \textit{arXiv}, 2018, arXiv:1804.02767.
\bibitem{SSD} W. Liu, D. Anguelov, D. Erhan, C. Szegedy, S. Reed, C. Y. Fu, A.C. Berg, “SSD: Single Shot MultiBox Detector,’’ \textit{The 14th European Conference on Computer Vision (ECCV2016)}, vol. 9905, 2016, pp. 21–37.
\bibitem{Shermayer} J. Shermeyer and A. V. Etten,“The Effects of Super-Resolution on Object Detection Performance in Satellite Imagery,’’ \textit{EarthVision 2019,IEEE}, 2019.
\bibitem{Koga} Y. Koga, H. Miyazaki, and R. Shibasaki, “A Method for Vehicle Detection in High-Resolution Satellite Images that Uses a Region-Based Object Detector and Unsupervised Domain Adaptation,’’ \textit{Remote Sensing 2020}, https://doi.org/10.3390/rs12030575.
\bibitem{xin} Z. Xin, H. Liangxiu, H. Lianghao, and Z. Liang, “How Well Do Deep Learning-Based Methods for Land Cover Classification and Object Detection Perform on High Resolution Remote Sensing Imagery?’’ \textit{Remote Sensing 2020}, 2020, https://doi.org/10.3390/rs12030417.
\bibitem{delmarco} S. P. DelMarco, V. Tom, H. Webb, W. Snyder, C. Jarvis, D. Fay, “Shape-based ATR for wide-area processing of satellite imagery,’’ \textit{SPIE 10988, Automatic Target Recognition XXIX}, 2019, https://doi.org/10.1117/12.2518185.
\bibitem{yanan} Y. Yanan, L. Zezhong, R. Bohao, C. Jingyi, L. Sudi, and L. Fang, “Broad Area Target Search System for Ship Detection via Deep Convolutional Neural Network,’’ \textit{Remote Sensing 2019}, 2019, https://doi.org/10.3390/rs11171965.
\bibitem{b3} R. A. Marcum, C. H. Davis, G. J. Scott, and T. W. Nivin, “Rapid broad area search and detection of Chinese surface-to-air missile sites using deep convolutional neural networks,’’ \textit{Journal of Applied Remote Sensing}, Vol. 11, No. 4, 042614, 2017, DOI: 10.1117/1.JRS.11.042614.
\bibitem{resnet} K. He, X. Zhang, S. Ren, and J. Sun, “Deep residual learning for image recognition,’’ \textit{Proc. of the IEEE Conf. on Computer Vision and Pattern Recognition}, pg. 770 –778, 2016, DOI: 10.1109/CVPR.2016.90.
\bibitem{b4} B. Zoph, V. Vasudevan, J. Shlens, and Q. Le, “Learning Transferable Architectures for Scalable Image Recognition,’’ \textit{CVPR 2018}, pg. 8697-8710, DOI: 10.1109/CVPR.2018.00907. 
\bibitem{xception} F. Chollet, "Xception: Deep Learning with Depthwise Separable Convolutions," \textit{2017 IEEE Conference on Computer Vision and Pattern Recognition (CVPR)}, pg. 1800-1807, doi: 10.1109/CVPR.2017.195.
\bibitem{proxylessnas} H. Cai,  L. Zhu,  and S. Han,  “ProxylessNAS: Direct neural architecture search on target task and hardware,”  \textit{International Conference on Learning Representations}, 2019, [Online]. Available:https://openreview.net/forum?id=HylVB3AqYm.
\bibitem{efficientnet} M.Tan, Q. V. Le, "EfficientNet: Rethinking Model Scaling for Convolutional Neural Networks", \textit{International Conference on Machine Learning}, 2019.
\bibitem{imagenet}J. Deng, W. Dong, R. Socher, L. J. Li, K. Li, and L. Fei-Fei,     “ImageNet: A Large-Scale Hierarchical Image Database,’’ \textit{2009 conference on Computer Vision and Pattern Recognition}, 2009,  DOI: 10.1109/CVPR.2009.5206848.
\bibitem{adam}D. P. Kingma and J. L. Ba, ‘‘Adam: A method for stochastic optimization,’’ Dec. 2014, arXiv:1412.6980. 
\bibitem{b5} A. B. Cannaday II, R. L. Chastain, J. A. Hurt, C. H. Davis, G. J. Scott and A. J. Maltenfort, “Decision-Level Fusion of DNN Outputs for Improving Feature Detection Performance on Large-Scale Remote Sensing Image Datasets,’’ \textit{2019 IEEE International Conference on Big Data (Big Data)}, Los Angeles, CA, USA, 2019, pp. 5428-5436, DOI: 10.1109/BigData47090.2019.9006502.
\bibitem{b6} N. Ye, K. M. A. Chai, W. S. Lee, H.L Chieu “Optimizing F-measures: a tale of two approaches,’’ \textit{Proceedings of the International Conference on Machine Learning}, 2012.
\bibitem{b7} D. D. Lewis, “Evaluating and optimizing autonomous text classification systems,’’ \textit{InSIGIR}, 1995, pp. 246–254.
\bibitem{b24} J. Jang, “ANFIS Adaptive-Network-based Fuzzy Inference System. Systems, Man and Cybernetics“, \textit{IEEE Transactions on. 23}. 1993, pp. 665 - 685, 10.1109/21.256541.
\bibitem{b22} A. Abraham, “Adaptation of Fuzzy Inference System Using Neural Learning,’’ in Nedjah, Nadia; de Macedo Mourelle, Luiza (eds.), \textit{Fuzzy Systems Engineering: Theory and Practice, Studies in Fuzziness and Soft Computing, 181, Germany: Springer Verlag}, 2005, pp. 53–83.
\bibitem{b23} D. Karaboga, and E. Kaya, “Adaptive network based fuzzy inference system (ANFIS) training approaches: a comprehensive survey,“ \textit{Artificial Intelligence Review}, 2018, DOI: 10.1007/s10462-017-9610-2.
\bibitem{b25} B. Ruprecht, C. Veal, B. Murray,M. Islam, D. Anderson, F. Petry, J. Keller, G. Scott, and C. Davis, “Fuzzy logic-based fusion of deep learners in remote sensing,’’ \textit{IEEE International Conference on Fuzzy Systems (FUZZ-IEEE)}, 2019.
\bibitem{b27} B. Ruprecht, C. Veal, A. Cannaday, D. Anderson, F. Petry, J. Keller, G. Scott, C.Davis, C. Northworthy, K. Nock, and E. Glimour, “Are neural fuzzy logic systems really explainable and interpretable?,’’ \textit{SPIE Security and Defense}, 2020.
\bibitem{b28} B. Ruprecht, C. Veal, B. Murray,M. Islam, D. Anderson, F. Petry, J. Keller, G. Scott, and C. Davis, “Possibilistic Clustering Enabled Neuro Fuzzy Logic,’’ under review, \textit{World Congress on Computational Intelligence}, 2020.

 
\end{thebibliography}
\end{document}